%% file: main.tex
\documentclass{svproc}
\usepackage{url}

\usepackage{amsmath}
\usepackage{amssymb}
\usepackage{algorithm} 
\usepackage[noend]{algpseudocode}
\usepackage{dirtytalk}
\usepackage{booktabs}
\usepackage{bm}
\usepackage{subcaption}
\usepackage{arydshln}
\usepackage{tikz}
\usepackage{tikz-3dplot}
\usepackage{hyperref}
\usetikzlibrary{shapes,arrows,positioning,calc,patterns,fit,shadows, decorations.pathmorphing, decorations.pathreplacing, 3d}

\algrenewcommand\algorithmicwhile{\textbf{Each agent $i$ does:}}
\algrenewcommand\algorithmicfor{\textbf{For}}
\algrenewcommand\algorithmicdo{}
\newcommand\norm[1]{\left\lVert#1\right\rVert}

\begin{document}
\mainmatter              
\title{Generalizing Differentially Private Decentralized Deep Learning with Multi-Agent Consensus}
\titlerunning{Differentially Private Decentralized Deep Learning}  
%
\author{Jasmine Bayrooti \and Zhan Gao \and Amanda Prorok}
\authorrunning{Jasmine Bayrooti et al.} 
%
\tocauthor{Jasmine Bayrooti, Zhan Gao, and Amanda Prorok}
\institute{University of Cambridge, Cambridge, UK,\\
\email{\{jgb52, zg292, asp45\}@cam.ac.uk}}

\maketitle              

\begin{abstract}
Cooperative decentralized learning relies on direct information exchange between communicating agents, each with access to locally available datasets. The goal is to agree on model parameters that are optimal over all data. However, sharing parameters with untrustworthy neighbors can incur privacy risks by leaking exploitable information. To enable trustworthy cooperative learning, we propose a framework that embeds \textit{differential privacy} into decentralized deep learning and secures each agent's local dataset during and after cooperative training. We prove convergence guarantees for algorithms derived from this framework and demonstrate its practical utility when applied to subgradient and ADMM decentralized approaches, finding accuracies approaching the centralized baseline while ensuring individual data samples are resilient to inference attacks. Furthermore, we study the relationships between accuracy, privacy budget, and networks' graph properties on collaborative classification tasks, discovering a useful invariance to the communication graph structure beyond a threshold.
\end{abstract}
%
\section{Introduction}


While centralized machine learning has led to valuable breakthroughs, decentralized learning offers advantages such as scalability and improved data privacy that are crucial for certain scenarios ~\cite{49232,NIPS2017_f7552665}. Specifically, when multiple agents (i.e., nodes in a graph) share an objective and have complementary local datasets that cannot be centralized due to scaling or confidentiality constraints, agents must cooperate to learn the best model in a \textit{decentralized} manner. As an example, this is pertinent to healthcare since hospitals desire to collaborate to train models that yield high-accuracy diagnoses, but are legally prohibited from aggregating patients' data into a centralized, shared training set~\cite{warnat2021swarm,rajkomar2019machine}.

Most research addressing data privacy in distributed systems uses federated learning, where a central server orchestrates training among participating agents, keeping data local to each agent~\cite{pmlr-v54-mcmahan17a,MLSYS2019_bd686fd6,49232}. This respects data confidentiality by avoiding transmission of raw data, however the central server remains a single point of vulnerability and direct parameter passing enables a determined attacker to infer characteristics of local training datasets ~\cite{8835245,9833677,10.1145/2810103.2813677,yeom2018privacy}. For instance, an attacker can perform membership inference attacks to deduce whether an agent has a given data sample in its private dataset by inspecting parameter values~\cite{8835245}. Therefore, federated learning is often used with differential privacy~\cite{10.1561/0400000042}, which introduces calibrated noise to mask the effects of individual data samples~\cite{NEURIPS2018_21ce6891,yang2023privatefl,stevens2022efficient}. Still, federated learning can be limited by computational bottlenecks, communication overhead, and low fault tolerance~\cite{NIPS2017_f7552665}. On the other hand, fully decentralized approaches distribute not only data-driven local updates but also system-wide learning, resulting in improved scalability. Decentralized agents, each with a private dataset and local model, learn by updating local parameters and exchanging information with neighbors, allowing for large-scale information dissemination without central components~\cite{ram2009asynchronous,yuan2016convergence}. Yet this local nature of interactions increases the attack surface, hence escalating the risk of white-box inference attacks in fully decentralized learning ~\cite{10179291}.

To address this, we propose a general framework for incorporating differential privacy~\cite{10.1561/0400000042}, the standard for preserving data privacy with formal guarantees, into first order, peer-to-peer decentralized algorithms by extending the single-agent DP-SGD approach~\cite{Abadi_2016}. We apply our framework to useful subgradient~\cite{pu2020distributed,NIPS2017_f7552665,DBLP:journals/corr/abs-2010-11166,NIPS2017_a74c3bae} and Alternating Direction Method of Multipliers (ADMM)~\cite{DBLP:journals/corr/abs-2109-08665} based methods to obtain algorithms that ensure differential privacy for each agent's local dataset on every training step by privatizing parameter exchanges. This fundamentally enables \textit{trustworthy cooperative learning} by guarding against the consequences of agents' potentially malicious inquisitiveness about other agents' parameters and local data, thereby unlocking potential for large-scale applications in privacy sensitive domains such as for drone delivery \cite{SHEIKABDULLAH20241}, disease diagnosis~\cite{warnat2021swarm}, financial fraud detection~\cite{WEST201647}, and public health surveillance~\cite{doi:10.1146/annurev-publhealth-031816-044348}.

While introducing differential privacy into decentralized learning seems natural, the key challenge is in controlling local noise to maintain differential privacy while still allowing convergence. In federated learning with differential privacy, the central server routinely aggregates across all agents' perturbed information~\cite{NEURIPS2018_21ce6891}. In contrast, injecting noise at the agent-level in fully decentralized settings results in patterns of noisy parameter aggregations for subsets of agents in the network. These patterns are difficult to track due to their dependence on parameter updates over time and the underlying communication graph, which is often dynamic in applications such as robotics~\cite{DBLP:journals/corr/abs-2109-08665}. In this work, under mild assumptions, we prove convergence guarantees when applying algorithms derived from our framework for arbitrary learning objectives. Our key insight is to craft the framework at a high-level in a way that maintains the convergence properties of the base algorithm. 

\textbf{Contributions.} This paper presents the following principal contributions:

\begin{itemize}
\item We present a general decentralized learning framework that yields agent-sample-level privacy guarantees that apply to any deep learning objective function on every step of cooperative training;
\item We formally prove our framework's bounded impact on convergence as algorithms derived from our framework, enhanced with differential privacy, inherit convergence properties from their base counterparts (Theorem \ref{thm:general_convergence});
\item When applied to select subgradient and ADMM based decentralized algorithms, namely DSGD, DSGT, and DiNNO, we showcase our framework's ability to achieve high accuracies (previously believed infeasible~\cite{cyffers2022muffliato}) relative to centralized DP-SGD and provide our open-sourced implementation \href{https://github.com/jbayrooti/dp-dec-learning}{here};
\item We experimentally discover a significant degree of invariance to agent interaction topology (i.e., graph structure) among the decentralized differentially private learning algorithms across various privacy budgets, which is a valuable insight that allows relaxation of the communication graph connectivity in bandwidth-limited situations without compromising performance; and,
\item Using membership inference attacks, we confirm that our framework equips these derived algorithms with practical levels of data privacy that are consistent with the privacy budget used during training. 
\end{itemize}

\section{Related Work}


In this section, we contextualize our contributions with a review of related work on decentralized optimization, differential privacy, and their intersection.

\noindent \textbf{Decentralized Optimization.} Diverse approaches to decentralized optimization can be categorized as subgradient or ADMM based methods~\cite{tsitsiklis,1638541,YANG2019278,DBLP:journals/corr/abs-2109-08665,4749425}. Subgradient approaches reach consensus by interleaving local gradient descent steps with averaging over values from neighboring agents~\cite{NIPS2017_f7552665,pu2020distributed}. Works have studied how factors such as network topology, objective convexity, and step size impact the rate of convergence~\cite{yuan2016convergence,6483403,NEURIPS2021_5f25fbe1,NIPS2017_a74c3bae,7398129,NEURIPS2020_f1ea154c} and have found that subgradient algorithms are well-suited for non-smooth and non-convex optimization problems \cite{ZHU2010322,doi:10.1137/08073038X,4749425,pu2020distributed,NIPS2017_f7552665}.
While subgradient approaches enforce implicit consensus via parameter averaging, ADMM methods facilitate constrained optimization, thus optimizing for both consensus and performance~\cite{6425904,6760448,8186925,5499155,10.1561/2400000003}.

\noindent \textbf{Differential Privacy.} The DP-SGD algorithm~\cite{Abadi_2016} provides privacy guarantees quantified by the moments accountant and is popular in centralized and federated learning settings~\cite{de2022unlocking,9069945,NEURIPS2018_21ce6891,10.1145/3491254}. PATE~\cite{papernot2018scalable} and DP-SRM~\cite{wang2023efficient} are alternative centralized approaches that offer improvements to DP-SGD's utility by relaxing the privacy guarantees. Although our contributions easily generalize to these other strategies, we focus on DP-SGD for stronger privacy protection.

\noindent \textbf{Differentially Private Decentralized Learning.} While differentially private federated learning focuses on privacy during aggregation at the central server~\cite{NEURIPS2018_21ce6891,pmlr-v202-murata23b}, fully decentralized differentially private learning must ensure privacy at the individual agent level. Some works \cite{xuan2023gradienttracking,10115431,8619119} present such approaches, although their methods protect agents' objective functions rather than agents' datasets. A recent study \cite{WANG2023110858} only bounds the privacy loss of a single training iteration and evaluates on non-standard tasks for differential privacy. Naive composition leads to an overestimation of the privacy loss~\cite{7563366,8049725}, hence it is crucial to analyze the privacy budget over all training iterations as we do. The LEASGD~\cite{cheng2019towards}, SDM-DSGD~\cite{zhang2020private}, and Q-DPSGD2~\cite{Ding_Liang_Bi_Pan_2021} algorithms are targeted towards communication efficiency, and hence involve exchanging quantized messages between agents. Aside from the sparsified messages, these algorithms fall under the umbrella of our framework. Finally, the Muffliato algorithm~\cite{cyffers2022muffliato} targets situations where agents require stronger privacy guarantees with respect to distant peers. Instead, our approach equips all agents with equal levels of privacy and makes no assumption of trust between any agents. Moreover, we are the first to show convergence for general, cumulative differentially private decentralized learning and demonstrate practical data privacy against membership inference attacks unlike all these works.


\section{Problem Statement}
\label{sec:problem}

In this section, we describe our problem setup and threat model.

\noindent \textbf{Setup.} Consider a cooperative deep learning problem involving $N$ agents, each with access to a private dataset and operating in an undirected, connected communication graph $\mathcal{G} = (\mathcal{V}, \mathcal{E})$. Let $\mathcal{D}_i$ be the subset of data that agent $i$ can access and $l(\cdot)$ be the objective function (same for all agents). We aim to optimize a neural network parameterized by the weights $\theta \in \mathbb{R}^d$ over the aggregate dataset $\mathcal{D}=\cup_{i \in \mathcal{V}} \mathcal{D}_i$, where agent $i$ stores its local estimate of the network weights $\theta_i$. We desire the agents to reach agreement on optimal network parameters $\theta^*$ after training. This distributed optimization problem can be formulated as:
\begin{align*}
\theta^* = \min_{\theta \in \mathbb{R}^d} \sum_{i=1}^N l(\theta;\mathcal{D}_i) = \min \sum_{i=1}^N l(\theta_i;\mathcal{D}_i) \text{\enspace s.t. $\theta_1= \dots =\theta_N$.}
\label{eq:problem}
\end{align*}

\noindent \textbf{Threat Model.} An attacker aims to infer or reconstruct sensitive information about agents’ local training data such as memberships or distributions. This attacker is an untrustworthy agent in the network that may passively exploit knowledge of parameter updates and model architecture (i.e., honest-but-curious).

\noindent \textbf{Communication.} Agents $i,j \in \mathcal{V}$ can exchange information if they are one-hop neighbors in the graph. Let $\mathcal{N}_i = \{i\} \cup \{j \in \mathcal{V} \mid (i,j) \in \mathcal{E}\}$ denote the local neighborhood of agent $i$. We parameterize communication links with a symmetric, doubly stochastic mixing matrix $\mathcal{W} \in \mathbb{R}^{N} \times \mathbb{R}^{N}$ with non-negative entries where $w_{ij} = 0$ if and only if nodes $i$ and $j$ are not connected. In this work, we assume communication is synchronous and securely encrypted so that a passive interceptor cannot infer meaningful qualities.

\noindent We additionally assume that any number of agents may be untrustworthy and all agents communicate their true values, otherwise agents may not reach consensus on model parameters. This threat model is a necessary first step towards stronger threat models that may address active and colluding attackers.

\section{Background}

We proceed by defining differential privacy and key decentralized algorithms.

\subsection{Differential Privacy}

Differential privacy ensures that, if two datasets differ by one individual data sample, the output of an algorithm does not reveal whether that sample was used. We specify such differences with the adjacency relation $Adj(\cdot, \cdot)$.

\begin{definition}[Adjacent Datasets~\cite{7798915}] Datasets $\mathcal{D}=\{(x,y)_i\}_{i=1}^n$ and \newline $\mathcal{D}'=\{(x',y')_i\}_{i=1}^n$ are adjacent if there exists an index $j \in [1,n]$ where $(x,y)_i=(x',y')_i$ for all $i \neq j$.
\label{def:adj}
\end{definition}
\noindent This allows us to formally introduce $(\epsilon,\delta)$-differential privacy.

\begin{definition}[Differential Privacy~\cite{NEURIPS2020_fc4ddc15}]
A mechanism $\mathcal{M}: \mathcal{X} \times \mathcal{Y} \rightarrow \mathcal{R}$ is $(\epsilon, \delta)$-differentially private if, for any adjacent datasets $\mathcal{D},\mathcal{D}'$ and every set of outputs $\mathcal{O} \subseteq \mathcal{R}$, the following holds:
\begin{equation}
\mathbb{P}[\mathcal{M}(\mathcal{D}) \in O] \leq e^{\epsilon}\mathbb{P}[\mathcal{M}(\mathcal{D}') \in O] + \delta.
\end{equation}
\end{definition}

\noindent The privacy budget $\epsilon$ bounds the log-likelihood ratio of obtaining the same outcome when running the algorithm on adjacent datasets, $\delta$ bounds the occurrence of outputs that violate the privacy limit.

\noindent \textbf{DP-SGD}~\cite{Abadi_2016} is a differentially private, single-agent, deep learning algorithm that performs stochastic gradient descent with a few modifications. Namely, samples are independently selected uniformly, per-sample gradients are clipped to a maximal norm $C$, and Gaussian noise with variance proportional to $C$ is added to gradients. Abadi et al.~\cite{Abadi_2016} uses the moments accountant to select the noise variance $\sigma$ so that DP-SGD provably satisfies a cumulative $(\epsilon,\delta)$-DP guarantee. In our implementation, we use the sampled Gaussian mechanism~\cite{mironov2019renyi}, which is closely related to the moments accountant, for the same purpose.

\subsection{Decentralized Optimization}
\label{sec:dist_ml}

We describe three diverse decentralized algorithms to which we will apply our framework. These approaches exhibit core similarities with other distributed algorithms and thus showcase generalizable ways to apply our framework.

DSGD~\cite{NIPS2017_f7552665} extends SGD to the distributed setting with the following update for agent $i$:
\begin{equation}
\label{eq:dsgd_update}
\theta^{k+1}_i =  \sum_{j \in \mathcal{N}_i} w_{ij} \theta^k_j - \eta^k \nabla l(\theta^k_i;\mathcal{D}_i),
\end{equation}
where $\eta^k$ is the learning rate at the $k$th iteration. For strongly convex objectives, DSGD converges to a neighborhood of the global optimum~\cite{NIPS2017_a74c3bae,pmlr-v119-koloskova20a} and is affected by non-IID data among agents~\cite{pmlr-v119-koloskova20a}.

DSGT~\cite{pu2020distributed} is a generic gradient tracking algorithm with updates:
\begin{equation}
y^{k+1}_i = \nabla l(\theta^{k}_i;\mathcal{D}_i) + \sum_{j \in \mathcal{N}_i} w_{ij}y_j^k - \nabla l(\theta^{k-1}_i;\mathcal{D}_i)
\label{eq:dsgt_y_update}
\end{equation}
\begin{equation}
\theta^{k+1}_i = \sum_{j \in \mathcal{N}_i} w_{ij}\left(\theta^k_j - \eta^k y_j^{k+1}\right).
\label{eq:dsgt_theta_update}
\end{equation}
where $y_i^k$ tracks the per-agent estimate of the joint loss gradient. Gradient tracking methods mitigate the impact of non-IID data distributions~\cite{7398129,xuan2023gradienttracking,6930814} and share convergence rates with mini-batch SGD for strongly convex objectives~\cite{NEURIPS2021_5f25fbe1}.

Finally, DiNNO~\cite{DBLP:journals/corr/abs-2109-08665} is a consensus ADMM-based algorithm that optimizes a dual variable $y_i^k$ to enforce inter-agent agreement and primal variables $\theta_i^k$ to minimize the joint loss. These equations describe the updates:
\begin{equation}
y_i^{k+1}=y_i^k+\rho \sum_{j \in \mathcal{N}_i}(\theta_i^k-\theta_j^k) 
\label{eq:primal_ascent}
\end{equation}
\begin{equation}
\theta_i^{k+1} = \text{argmin}_{\theta} l(\theta;\mathcal{D}_i)+\theta^T y_i^{k+1}+\rho \sum_{j \in \mathcal{N}_i}\norm{\theta - \frac{\theta_i^k+\theta_j^k}{2}}_2^2
\label{eq:primal_optimization}
\end{equation}
where $\rho$ acts as the step size for gradient ascent of the dual variable in Equation \ref{eq:primal_ascent} and weights the quadratic regularization term in Equation \ref{eq:primal_optimization}. For strongly convex objectives, \cite{DBLP:journals/corr/abs-2109-08665} shows that DiNNO converges to the unique global solution.


\section{General Decentralized Differentially Private Learning}
\label{sec:methods}
In decentralized algorithms, agents share their local parameters with neighbors, thus incurring privacy risk since training samples can leave a discernible trace on gradients and, transitively, parameters~\cite{8835245}. To address this, we describe a general framework for integrating differential privacy into decentralized learning.

\subsection{Framework for Decentralized Differentially Private Learning}

In first-order decentralized approaches, agent $i$ updates its local parameters as a function of aggregated information from its neighbors $\mathcal{N}_i$ and the first-order gradient $\nabla l(\theta_i;\mathcal{D}_i)$ of the objective function with respect to $\mathcal{D}_i$. Our framework leverages this commonality to equip each agent with \textit{local differential privacy} by adding calibrated noise to the gradient term, in a generalization of DP-SGD. Specifically, on each training step, agent $i$ uniformly selects $L$ data samples from $\mathcal{D}_i$ to form a \textit{lot} $\mathcal{L}_i$. Then agent $i$ computes the gradient of the objective function and clips the $l_2$ norm of per-sample gradients with $\min \{1, \frac{C}{||v||_2}\} \cdot v \in \mathbb{R}^d$ for $v \in \mathcal{L}_i$. Clipping per-sample gradients is key for bounding the influence of individual samples on the overall lot gradient~\cite{Abadi_2016}. Agent $i$ then aggregates the clipped gradients in the lot, adds Gaussian noise, and rescales the lot gradient to obtain $\tilde{G}$. We utilize the sampled Gaussian mechanism~\cite{mironov2019renyi} to select the suitable standard deviation $\sigma$ when adding Gaussian noise to gradients and ensure cumulative $(\epsilon,\delta)$-DP for each agent. Note that differential privacy for the gradients transfers to the resulting model as post processing maintains the differential privacy property. The inter-agent communication step and parameter update subroutine $\Psi$ are carried out according to the base algorithm using the new lot gradient $\tilde{G}$. We outline this process in Algorithm \ref{alg:framework}.

\begin{algorithm}
	\caption{General Decentralized DP Learning}
    \label{alg:framework}
	\begin{algorithmic}[1]
    \State \textbf{Require:} $\mathcal{G}$, $\mathcal{D}$, $l(\cdot)$, $L$, $C$, $\Psi$, $\eta$, $\sigma$, $K$
    \For {$k = 1,2,\dots,K$}
        \State \textbf{Communicate}: send values to neighbors $\mathcal{N}_i$
        \While \Comment{In parallel}
            \State Take random samples $\mathcal{L}_i^k$ with probability $L/|\mathcal{D}_i|$
            \State $\xi \sim N(0, I_d)$ \Comment{Draw Gaussian sample}
            \State $\tilde{G}(\theta_i^k;\mathcal{D}_i) = \frac{1}{L} \sum_{v \in \mathcal{L}^k_i} \texttt{clip}_C \left(\nabla l(\theta^k_i;v)\right) + \frac{\sigma C}{L}\xi$ \Comment{Clip and add noise}
            \State $\theta_i^{k+1} = \Psi(\tilde{G}(\theta_i^k;\mathcal{D}_i), \eta, \dots)$ \Comment{Update parameters}
        \EndWhile
    \EndFor
	\end{algorithmic}
\end{algorithm}

\subsection{Convergence Analysis}
\label{sec:conv_anal}

We analyze convergence properties for algorithms derived with our framework, showing that differentially private decentralized algorithms share convergence properties with their base decentralized algorithms under mild conditions. Let $\theta \in \mathbb{R}^{Nd}$ be a vector concatenating local variables $\{\theta_i\}_{i=1}^N$ and $M(\theta)$ be a function of $\theta$ for which the base algorithm converges.

\begin{theorem}[Linear Convergence Theorem]
Assume that $M(\theta)$ of the base decentralized algorithm is Lipschitz w.r.t. a constant $C_L$, i.e., $\|M(\theta) - M(\hat{\theta})\|_2 \le C_L \|\theta - \hat{\theta}\|_2$ for any $\theta$ and $\hat{\theta}$. If $M(\theta)$ of the base decentralized algorithm converges to the optimal solution $M(\theta^*)$ at a linear rate, then $M(\theta)$ of the differentially private decentralized algorithm also converges to the optimal solution $M(\theta^*)$ at a linear rate within an error neighborhood on the order of $\mathcal{O}(\sigma)$.
\label{thm:lin_convergence}
\end{theorem}

\proof{
Let $\tilde{\theta}^{(k)}$ denote the differentially private algorithm's noisy variable vector at iteration $k$, $\rho \in (0,1)$ be the contraction factor of the linear convergence, $C_\sigma$ be a scaling constant depending on the specific update step $\Psi(\cdot)$, and $\xi^{(k)}$ be the sampled noise at iteration $k$. By induction, one can verify that
\begin{equation}
\label{eq:ind_hyp}
    \| M\big(\tilde{\theta}^{(k)}) - M(\theta^*) \| \le \rho^{k} \| M\big(\theta^{(0)}\big) - M(\theta^*) \| + C_L C_\sigma \sum_{i=0}^{k} \rho^i \eta^{(k-i)} \|\xi^{(i)}\|
\end{equation}
holds for all iterations $k$. Taking the expectation over Equation \ref{eq:ind_hyp} and simplifying terms (bounding the sum over $\rho$ and letting $\eta^{(k)} \le C_\eta$), we have
\begin{equation}
     \mathbb{E} \big[ \| M\big(\tilde{\theta}^{(k)}) - M(\theta^*) \| \big] \le \rho^k \mathbb{E} \big[ \| M\big(\theta^{(0)}\big) - M(\theta^*) \| \big] + \frac{C_L C_\sigma C_\eta }{1-\rho} D, \nonumber
\end{equation}
where $D = \mathbb{E} \big[ \|\xi\| \big] = \mathcal{O}(\sigma)$. As desired, this shows that
\begin{equation}
     \lim_{k \to \infty}\mathbb{E} \big[ \| M\big(\tilde{\theta}^k) - M(\theta^*) \| \big] \le \mathcal{O}(\sigma). \nonumber
\end{equation}
}

\noindent The base algorithms DSGD, DSGT, and DiNNO converge linearly when the objective function is strongly convex~\cite{NIPS2017_a74c3bae,pu2020distributed,DBLP:journals/corr/abs-2109-08665}. For other cases, the following theorem generalizes our convergence analysis with additional assumptions.

\begin{theorem}[General Convergence Theorem]
Assume the gradient function $\nabla l(\cdot)$ is Lipschitz w.r.t the variable $\theta$, the update step of the base algorithm $\Psi(\cdot)$ is Lipschitz w.r.t. the gradient $G$ and normalized Lipschitz w.r.t. the variable $\theta$, the step size satisfies $\eta^k \le \frac{\hat{\eta}^k}{C k}$ at each iteration $k$ (where $\hat{\eta}^k \le 1$ is any sequence of values with a finite sum $\sum_{k=1}^\infty \hat{\eta}^k \le \infty$), and $C$ is a constant depending on the Lipschitz properties of $\nabla l(\cdot)$ and $\Psi$. If $\theta$ of the base decentralized algorithm converges to the optimal solution $\theta^*$, then $M(\theta)$ of the differentially private decentralized algorithm also converges to the optimal solution $M(\theta^*)$ in an error neighborhood of $\mathcal{O}(\sigma)$.
\label{thm:general_convergence}
\end{theorem}

\proof{
Let $\tilde{\theta}^{(k)}$ and $\theta^{(k)}$ denote the differentially private and base algorithms' variable vectors respectively. We observe that $\tilde{\theta}^{(k)}$ deviates from $\theta^{(k)}$ with the accumulated noise from previous iterations upper bounded with
\begin{align}\label{eq:0}
\mathbb{E}[\|\tilde{\theta}^{(k+1)} - \theta^{(k+1)}\|] &\le C_{\sigma} \eta^{k+1} \big(\mathbb{E}[\|\tilde{\theta}^{(k)} - \theta^{(k)}\|] + \mathbb{E}[\|\xi^{(k+1)}\|] \big) + \mathbb{E}[\|\tilde{\theta}^{(k)} - \theta^{(k)}\|] \nonumber
\end{align}
where $C_{\sigma}$ depends on the Lipschitz constants of the gradient function and the update step of the base algorithm. The first term in the bound comes from the gradient deviation and the second from the variable deviation. We can further bound the difference using induction and the condition that $\eta^k \le \frac{\hat{\eta}^k}{C k}$, with
\begin{equation}\label{eq:1}
\mathbb{E}[\|\tilde{\theta}^{(k)} - \theta^{(k)}\|] \le \sum_{\kappa=1}^{k} \hat{\eta}^k \sigma. \nonumber
\end{equation}
Using this and the condition that there exists $C_\eta$ such that $\sum_{k=1}^\infty \hat{\eta}^k \le C_\eta > 0$, we find $\mathbb{E}[\|\tilde{\theta}^{(k)} - \theta^{(k)}\|] \le C_\eta \sigma$. Therefore, we conclude the proof that
\begin{equation}
\lim_{k \to \infty} \mathbb{E}[\|M(\tilde{\theta}^{(k+1)} - M(\theta^{(k+1)}\|] \le C_L C_\eta \sigma. \nonumber
\end{equation} 
}


\section{Applying the General Decentralized DP Framework}

We apply our framework to DSGD, DSGT, and DiNNO by utilizing the relevant subroutine $\Psi$ in Algorithm \ref{alg:framework} to develop their differentially private counterparts.

\noindent \textbf{DP-DSGD} entails sharing parameters $\theta_i^k$ with neighbors and updating parameters using a differentially private version of Equation \ref{eq:dsgd_update}. We describe the approach in Algorithm \ref{alg:DP-DSGD} and use the following natural definition for $\Psi$:

\begin{equation}
\Psi(\tilde{G}(\theta_i^k;\mathcal{D}_i), \eta, \{\theta_j^k\}_{j \in \mathcal{N}_i}, \mathcal{W}) = \sum_{j \in \mathcal{N}_i} w_{ij}\theta^k_j - \eta^k \tilde{G}(\theta_i^k;\mathcal{D}_i). \nonumber
\end{equation}

\begin{algorithm}[h]
	\caption{DP-DSGD}
    \label{alg:DP-DSGD}
	\begin{algorithmic}[1]
    \State \textbf{Require:} $\mathcal{G}$, $\mathcal{D}$, $l(\cdot)$, $\mathcal{W}$, $\theta_{\mathrm{initial}}$, $L$, $C$, $\eta$, $\sigma$, $K$
    \While {} \Comment{In parallel}
        \State $\theta_i^1 = \theta_{\mathrm{initial}}$ \Comment{Initialize parameters}
    \EndWhile
    \For {$k = 1,2,\dots,K$}
        \State \textbf{Communicate}: send $\theta^k_i$ to neighbors $\mathcal{N}_i$
        \While \Comment{In parallel}
            \State Take random samples $\mathcal{L}_i^k$ with probability $L/|\mathcal{D}_i|$
            \State $\xi \sim N(0, I_d)$ \Comment{Draw Gaussian sample}
            \State $\tilde{G}(\theta_i^k;\mathcal{D}_i) = \frac{1}{L} \sum_{v \in \mathcal{L}^k_i} \texttt{clip}_C \left(\nabla l(\theta^k_i;v)\right) + \frac{\sigma C}{L}\xi$ \Comment{Clip and add noise}
            \State $\theta_i^{k+1} = \sum_{j \in \mathcal{N}_i} w_{ij}\theta^k_j - \eta^k \tilde{G}(\theta_i^k;\mathcal{D}_i)$ \Comment{DP version of \eqref{eq:dsgd_update}}
        \EndWhile
    \EndFor
	\end{algorithmic} 
\end{algorithm}

\noindent DP-DSGD meets our convergence conditions since we assume $M(\theta)=\frac{1}{N} \sum_{i=1}^N \theta_i$, in accordance with the DSGD convergence analysis in \cite{NIPS2017_f7552665}, and this is Lipschitz.

\noindent \textbf{DP-DSGT} involves computing gradient estimates $y_i^k$ with a differentially private version of Equation \ref{eq:dsgt_y_update}. We use this to update parameters $\theta_i^k$ by taking an $\eta$-weighted step in the opposite direction of the estimated objective gradient, which makes up the $\Psi$ mechanism. The full DP-DSGT approach is given in Algorithm \ref{alg:DP-DSGT} and this meets the convergence conditions since we follow the DSGT convergence analysis~\cite{pu2020distributed} and let $M(\theta)=\frac{1}{N} \sum_{i=1}^N \theta_i$, which is Lipschitz.

\begin{algorithm}[t]
	\caption{DP-DSGT}
    \label{alg:DP-DSGT}
	\begin{algorithmic}[1]
    \State \textbf{Require:} $\mathcal{G}$, $\mathcal{D}$, $l(\cdot)$, $\mathcal{W}$, $\theta_{\mathrm{initial}}$, $L$, $C$, $\eta$, $\sigma$, $K$
    \While {} \Comment{In parallel}
        \State $\theta_i^1 = \theta_{\mathrm{initial}}$, $y_i^1 = 0$, $\tilde{G}^1(\theta_i^0;\mathcal{D}_i) = 0$ \Comment{Initialize values}
    \EndWhile
    \For {$k = 1,2,\dots,K$}
        \While {} \Comment{In parallel}
            \State Take random samples $\mathcal{L}_i^k$ with probability $L/|\mathcal{D}_i|$
            \State $\xi \sim N(0, I_d)$ \Comment{Draw Gaussian sample}
            \State $\tilde{G}^{k+1}(\theta_i^{k};\mathcal{D}_i) = \frac{1}{L} \sum_{v \in \mathcal{L}^k_i} \texttt{clip}_C \left(\nabla l(\theta^{k}_i;v)\right) + \frac{\sigma C}{L}\xi$ \Comment{Clip and add noise}
            \State $y^{k+1}_i = \tilde{G}^{k+1}(\theta_i^{k};\mathcal{D}_i) + \left(\sum_{j \in \mathcal{N}_i} w_{ij}y_j^k - \tilde{G}^{k}(\theta_i^{k-1};\mathcal{D}_i) \right)$ \Comment{DP version of \eqref{eq:dsgt_y_update}}
        \EndWhile
        \State \textbf{Communicate}: send $\theta^k_i$, $y^{k+1}_i$ to neighbors $\mathcal{N}_i$
        \While {} \Comment{In parallel}
            \State $\theta^{k+1}_i = \sum_{j \in \mathcal{N}_i} w_{ij}(\theta^k_j - \eta^k y_j^{k+1})$
        \EndWhile
    \EndFor
	\end{algorithmic} 
\end{algorithm}

\noindent \textbf{DP-DiNNO} involves repeatedly solving the optimization problem in Equation \ref{eq:primal_optimization}. Since this is challenging to do analytically when $l$ is non-convex, most ADMM approaches assume $l$ is convex and bound the update function's sensitivity using the closed form solution~\cite{DBLP:journals/corr/abs-2005-07890,7563366}. Instead, we define the procedure $\Psi$ with a nested loop and solve Equation \ref{eq:primal_optimization} iteratively as in \cite{DBLP:journals/corr/abs-2109-08665}, while injecting calibrated noise for differential privacy. Note that we do not need to clip or add noise to the other gradient terms in Equation \ref{eq:primal_optimization} since only $\nabla l(\theta_i;\mathcal{D}_i)$ relies on agent $i$'s private dataset $\mathcal{D}_i$. We outline DP-DiNNO in Algorithm \ref{alg:DP-DiNNO} and confirm it converges with the Lipschitz function $M(\theta)=\theta$, where $\theta$ concatenates the primal and dual variables, as is standard in ADMM convergence analysis~\cite{6731604}.

\begin{algorithm}[t]
	\caption{DP-DiNNO}
    \label{alg:DP-DiNNO}
	\begin{algorithmic}[1]
    \State \textbf{Require:} $\mathcal{G}$, $\mathcal{D}$, $l(\cdot)$, $\theta_{\mathrm{initial}}$, $L$, $C$, $\rho$, $\eta$, $\sigma$, $K$, $T$
    \While {} \Comment{In parallel}
        \State $\theta_i^1 = \theta_{\mathrm{initial}}$, $y_i^1 = 0$ \Comment{Initialize primal and dual variables}
    \EndWhile
    \For {$k=1,2,\ldots,K$}
        \State \textbf{Communicate}: send $\theta^k_i$ to neighbors $\mathcal{N}_i$
        \While {} \Comment{In parallel}
            \State $y_i^{k+1}=y_i^k+\rho \sum_{j \in \mathcal{N}_i}(\theta_i^k-\theta_j^k)$ \Comment{Update the dual}
            \State $\psi^1 = \theta_i^k$ \Comment{Warm start primal optimization}
            \For {$t=1,2,\ldots,T$}
                \State Take random samples $\mathcal{L}_i^k$ with probability $L/|\mathcal{D}_i|$
                \State $\xi \sim N(0, I_d)$ \Comment{Draw Gaussian sample}
                \State $\tilde{G}_{\mathrm{data}}(\psi^t;\mathcal{D}_i) = \frac{1}{L} \sum_{v \in \mathcal{L}_i^k} \texttt{clip}_C \left(\nabla l(\psi^t;v)\right) + \frac{\sigma C}{L}\xi$ \Comment{Clip and add noise}
                \State $\tilde{G}(\psi^t) = \tilde{G}_{\mathrm{data}}(\psi^t;\mathcal{D}_i) + \nabla \left((\psi^t)^T y_i^{k+1}\right) + \nabla \left(\rho \sum_{j \in \mathcal{N}_i}\norm{\psi^t - \frac{\theta_i^k+\theta_j^k}{2}}_2^2\right)$
                \State $\psi^{t+1} = \psi^t + \eta \tilde{G}(\psi^t)$ \Comment{Take optimizer step}
            \EndFor
            \State $\theta_i^{k+1} = \psi^T$ \Comment{Update the primal variable}
        \EndWhile
    \EndFor
	\end{algorithmic} 
\end{algorithm}


\section{Experiments}
\label{sec:exps}

We evaluate the algorithms derived from our framework on two distributed classification problems. In this section, we describe these problems, experimental setup, and discuss results showing the relationships between accuracy, privacy budget, algebraic graph connectivity, node centrality, and data distribution.

\subsection{Experimental Setup}

\textbf{Tasks.} We evaluate on tasks that mirror practical scenarios where agents gather unique samples by exploring diverse parts of the environment and learn from collective experiences. For instance, drones can learn how to identify objects they fly over from shared diverse experiences. MNIST and CIFAR-100 are natural datasets for evaluation since they are publically available and have been used to benchmark differentially private deep learning for single-agent tasks~\cite{Abadi_2016,10.5555/3361338.3361469,zhang2020private}. We construct each of $N$ agents' local datasets by assigning them equal sized, non-overlapping portions of the entire training dataset. All agents use the same validation set to measure performance.

\noindent \textbf{Distributing Data to Agents.} Since we observe that similarity of data samples in a local dataset influences consensus and accuracy, we define a metric to quantify the distribution of data for each agent as a function of class labels.
\begin{definition}[Data Distribution]
Let the matrix $A(t)$ have rows corresponding to agents and columns corresponding to classes in the dataset. Then define:
\begin{equation}
a_{ij}(t) = \begin{cases} \frac{1-t}{N} \text{\quad $i \neq j$ mod $N$} \\ 1 - \sum_{k \neq i} a_{kj} \text{\quad if $i = j$ mod $N$.} \end{cases}
\end{equation}
\end{definition}
\noindent We distribute data to agents according to $A(t)$ so that agent $i$ has the fractional amount $a_{ij}$ of data labeled with class $j$ in $\mathcal{D}_i$. Hence, $t=0$ signifies that each agent has an equal distribution of every class and $t=1$ indicates that each agent has complete and exclusive access to specific classes.

\noindent \textbf{Training Details.} We implement DP-DSGD, DP-DSGT, and DP-DiNNO with PyTorch using the Opacus library~\cite{DBLP:journals/corr/abs-2109-12298} for differential privacy utilities and compare performance with the centrally-trained DP-SGD algorithm. To ensure a fair comparison, we use Ray's population-based tuning algorithm~\cite{ray} to select appropriate learning rates for each method and privacy budget, which we also hand-tune slightly. For all experiments, we use a uniform mixing matrix and $\delta = 10^{-5}$, gradient clipping threshold $C=1$ for DP-SGD training (as in \cite{de2022unlocking}), and $C=10$ for DP-DSGD, DP-DSGT, and DP-DiNNO (since these algorithms converge slower). We train MNIST experiments for 2,000 iterations using a shallow model (one convolutional layer and two linear layers) for each agent as done in \cite{DBLP:journals/corr/abs-2109-08665}. To achieve higher accuracies on CIFAR-100, we train for 20,000 iterations and use ResNet-9 models with batch norms replaced with group norms following the DP standard~\cite{de2022unlocking}. All other details are available in our repository \href{https://github.com/jbayrooti/dp-dec-learning}{here}.

\noindent \textbf{Computing Demands.} We simulate multi-agent training by independently training per-agent models simultaneously on a single GPU. Due to the large size of the ResNet-9 model, we limit our distributed CIFAR-100 experiments to $N=10$ to keep total GPU memory usage below our 40 GB capacity. Since it is superfluous to use more than $N=10$ agents for distributed MNIST classification because there are 10 classes total, for all experiments we consider a network with $N=10$ agents, each maintaining their own local datasets and distinct models and communicating over a connected graph.


\begin{figure*}[t]
  \centering
  \includegraphics[width=0.98\linewidth]{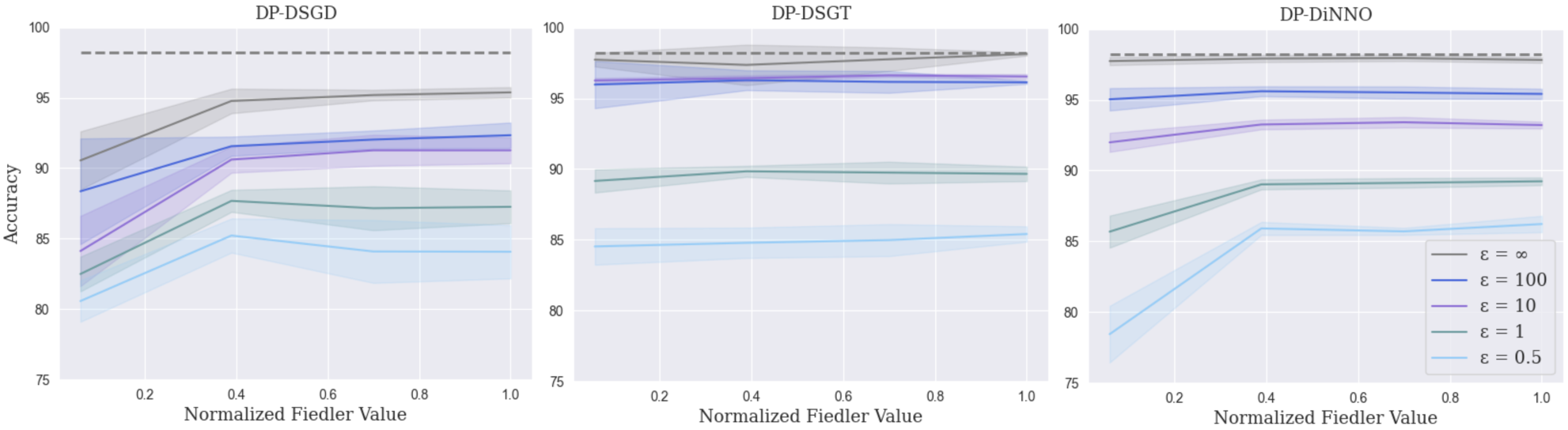}
  \setlength{\belowcaptionskip}{-3mm}
  \caption{All DP decentralized algorithms are relatively invariant to connectivity up to a threshold value near 0.4. The non-private accuracy from single-agent SGD is denoted with a dashed line.}
 \label{fig:connectivities}
\end{figure*}

\begin{figure*}[h]
  \centering
  \includegraphics[width=0.98\linewidth]{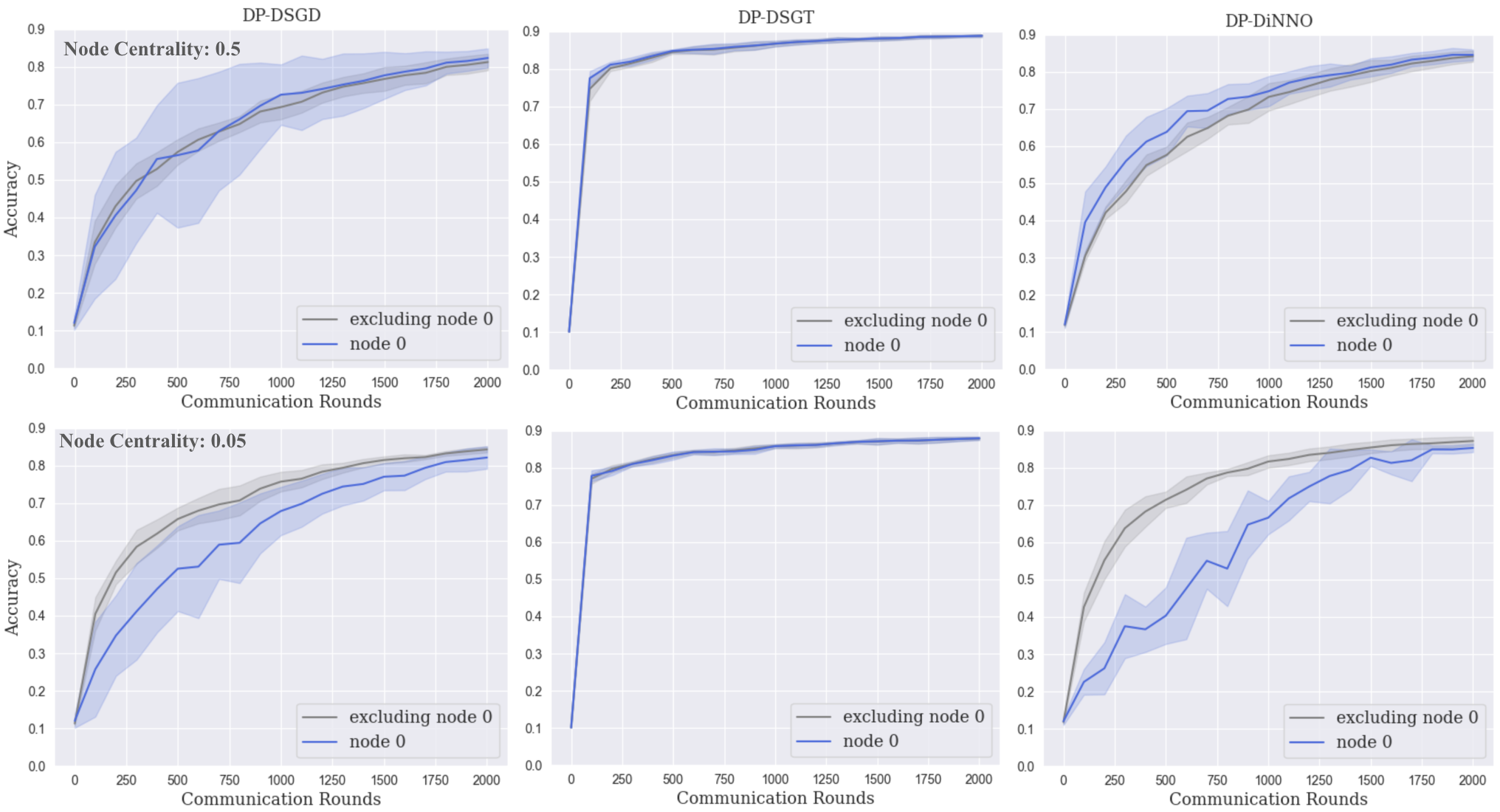}
  \setlength{\belowcaptionskip}{-3mm}
  \caption{DP-DiNNO is most susceptible to slow convergence when agents have low node centrality (bottom). DP-DSGD, DP-DSGT, and DP-DiNNO experience minor impacts on training for medium-to-high node centralities (top).}
 \label{fig:node_centrality}
\end{figure*}
\vspace{-0.5mm}

\subsection{Algebraic Connectivity}
\label{sec:alg_con}


The structure of the underlying communication graph influences consensus rates and performance for some decentralized algorithms~\cite{NEURIPS2021_74e1ed8b,song2022communication,koloskova2020unified}. To study this in the presence of differential privacy, we vary the privacy budget as well as the Fiedler value, which quantifies algebraic connectivity by measuring global connectedness of the graph. We train DP-DSGD, DP-DSGT, and DP-DiNNO on distributed MNIST with $t=1$ using randomly generated communication graphs of normalized Fiedler values close to one of four selected values (0.06, 0.39, 0.7, and 1) for five trials per datapoint. Our numerical findings confirm that centrally-trained DP-SGD always upper bounds the decentralized algorithms' accuracy by a few percentage points for each privacy budget. Nevertheless, we find that our algorithms can achieve strong performance on MNIST classification even for low privacy budgets, with DP-DSGT reaching accuracies around 3 percentage points of DP-SGD under cumulative $(1,10^{-5})$-DP. 

Furthermore, we isolate the effects of changing graph connectivity and privacy budgets in Figure \ref{fig:connectivities}. These findings indicate that DP-DSGT is invariant to graph connectivity, which likely stems from DP-DSGT's exchange of per-agent global gradient estimates in addition to model parameters. This can facilitate effective tracking of the global gradient, however requires exchanging twice the message size of DP-DSGD and DP-DiNNO, so may be infeasible for some applications. Figure \ref{fig:connectivities} also demonstrates that DP-DSGD and DP-DiNNO exhibit graph invariance for normalized Fiedler values above 0.4 and declining performance below 0.4. The consistent performance of all algorithms across graph connectivities above a normalized Fiedler value of 0.4 is significant because it implies we can relax the connectivity of the communication graph to a threshold in bandwidth-limited scenarios without compromising performance.

\subsection{Node Centrality}
\label{sec:node_centrality}

We continue studying the impact of the communication graph topology by honing in on a single node's structure and investigating how it influences that agent's rate of consensus. Algebraic connectivity is a graph-wide property, while node centrality quantifies an individual node's importance in a network. In our experiments, we use eigenvector node centrality, which characterizes importance based on connections to other significant nodes. We generate communication graphs so that node 0 has centrality close to 0.05 or 0.5 and all other structures are random. Training five trials of $(1, 10^{-5})$-DP-DSGD, DP-DSGT, and DP-DiNNO on distributed MNIST using these graphs yields validation accuracies depicted in Figure \ref{fig:node_centrality}. These results indicate that node 0's centrality significantly impacts consensus rates for DP-DSGD and DP-DiNNO, for which parameter consensus is slower with low node centrality. In contrast, we find that non-private DSGD and DiNNO are minimally affected by node centrality.

\subsection{Distribution of Data}
\label{sec:data_dist}

\begin{figure}[t]
  \centering
  \includegraphics[width=0.8\linewidth]{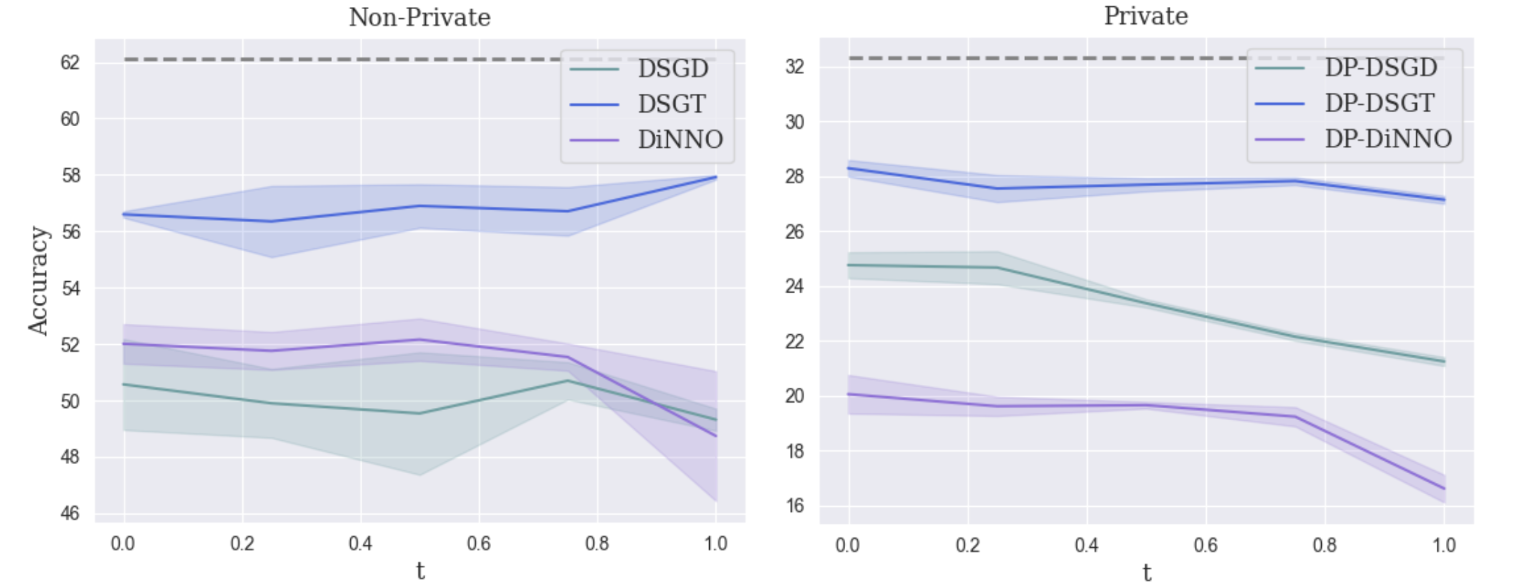}
  \setlength{\belowcaptionskip}{-3mm}
  \caption{DP-DSGT is relatively invariant to changes in data distributions in both non-private (left) and $(10, 10^{-5})$-differentially private (right) settings. DP-DiNNO and DP-DSGD experience more pronounced performance drops for less overlapping per-agent data distributions. Differential privacy induces a notable performance drop relative to non-private training, yet DP-DSGT remains within 6 percentage points of DP-SGD's accuracy (denoted with a dashed line).}
 \label{fig:data_split}
\end{figure}

Data heterogeneity poses another challenge in decentralized learning~\cite{NEURIPS2021_5f25fbe1,yang2023privatefl}. In this set of experiments, we study how the data distribution among agents impacts differentially private learning by varying $t$ on the more challenging distributed CIFAR-100 problem. We evaluate our non-private and ($10$, $10^{-5})$-DP algorithms on distributed CIFAR-100 using a range of $t$ values (0, 0.25, 0.5, 0.75, and 1) with a complete communication graph using three trials per datapoint. Figure \ref{fig:data_split} shows that our differentially private algorithms successfully generalize to harder tasks and deeper models. Most notably, DP-DSGT attains accuracies within 6 percentage points of DP-SGD across all five $t$ values. We additionally find that DP-DSGT maintains consistent performance regardless of the data distribution scheme, which agrees with results for non-private DSGT~\cite{7398129,xuan2023gradienttracking,6930814,NEURIPS2021_5f25fbe1}, and shows that incorporating differential privacy via our framework preserves this property of DSGT. This also demonstrates that DP-DSGT is a good choice for privacy-sensitive applications where agents own extremely heterogeneous data classes. On the other hand, DP-DSGD and DP-DiNNO benefit more from firsthand experience with each class since they drop off in performance as $t$ goes to 1.

\section{Membership Inference Attacks}

Finally, we validate that our framework equips decentralized algorithms with practical data privacy by evaluating their protection against membership inference attacks. We do this by adapting single-agent attacks~\cite{NEURIPS2020_fc4ddc15,de2022unlocking} to our multi-agent setting and empirically lower bounding our algorithms' privacy protection, which we compare with the nominal privacy guarantees. 

To do this, we select 100 images each of digits labeled 0, 1, and 2 from MNIST to form the local datasets for three agents. We let the aggregate dataset be $D$ and construct the \say{poisoned} dataset $D' = D \cup \{z\}$ where $z$ is a blank image labeled 0 (as in \cite{de2022unlocking}). We train 5,000 models on both $D$ and $D'$ using DP-DSGD, DP-DSGT, DP-DiNNO along with the centralized baseline DP-SGD for 100 iterations under $(1, 10^{-2})$-DP. We use 2,000 trained models to set the threshold for a simulated attacker to use when predicting whether or not $z$ is a training member given a trained model as input. We use the remaining training models to compute the attacker's true positive rate (TPR) and false positive rate (FPR), which satisfies $\frac{TPR - \delta}{FPR} < e^\epsilon$ for any $(\epsilon, \delta)$-DP algorithm~\cite{composition}. Taking the natural logarithm gives empirical lower bounds on $\epsilon$, which we report in Table \ref{tab:attack}. As the empirical $\epsilon$ values are lower than the nominal $\epsilon=1$, we find no violation of our $(\epsilon, \delta)$-DP guarantees. Note the gap in values is expected~\cite{9519424} and these lower bounds provide evidence that our differentially private algorithms offer levels of privacy that are unlikely to be weaker than we claim.
\vspace{-6mm}
\begin{table}[h]
  \caption{$\epsilon$ lower bounds on differential privacy guarantees.}
  \label{tab:attack}
  \centering
  \begin{tabular}{l@{\hspace{6mm}}l@{\hspace{6mm}}l@{\hspace{6mm}}l@{\hspace{6mm}}l}
  \toprule
   Method & DP-SGD & DP-DSGD & DP-DSGT & DP-DiNNO \\
  \midrule
  Empirical $\epsilon$ & 0.230 & 0.173 & 0.129 & 0.165 \\
  \bottomrule
  \end{tabular}
\end{table}
\vspace{-6mm}

\section{Conclusion}

In this paper, we proposed a framework incorporating differential privacy into arbitrary first-order decentralized learning algorithms to facilitate trustworthy cooperative deep learning. We proved that algorithms derived using our framework share convergence properties with their base algorithms. We applied this framework to DSGD, DSGT, and DiNNO and studied their properties with differential privacy under variable privacy budgets, algebraic connectivity, node centrality, and data distributions. Our findings revealed that these algorithms are unaffected by graph structure beyond a threshold, showing that connectivity can be reduced without sacrificing performance in bandwidth-limited scenarios. Ultimately, we showcased high-performing differentially private decentralized learning for challenging classification tasks, deep models, low privacy budgets, sparse graphs, and heterogeneous data distributions, all without assuming trust among agents and guarding against inference and reconstruction attacks. Expanding the scope of differential privacy in deep learning to encompass arbitrary first-order decentralized learning algorithms holds significant importance, and we anticipate that our work will lay the foundation for future advancements in this area.



\section{Acknowledgements}

J. Bayrooti is supported by a DeepMind scholarship. Z. Gao and A. Prorok are supported in part by European Research Council (ERC) Project 949940 (gAIa). We also thank Carl Henrik Ek, George Pappas, Alex Sablayrolles, Ajay Shankar, Alastair Beresford, Daniel Hugenroth, and the Cambridge security group.

\bibliographystyle{spmpsci} 
\bibliography{reference}

\newpage
\appendix

\section{Problem Motivation}

In this section, we provide additional motivation for our problem setup, described in Section \ref{sec:problem}. Consider the following real-world examples of such problems.

\begin{example}\label{ex:drones}
In a drone delivery system, a network of drones must deliver packages while learning the environment map during flight (e.g., to identify potential obstacles or land features), as depicted in Figure \ref{fig:hero}. Since location privacy is paramount to preserve the confidentiality of recipients' addresses, privacy-sensitive fully decentralized learning can achieve this on a larger scale than federated learning.
\end{example}

\begin{example}\label{ex:healthcare}
In the era of data-driven healthcare, hospitals can train models to enhance their diagnoses. However, stringent privacy regulations demand safeguarding patient data. To unlock the collective experiences across independent datasets without compromising privacy, data private fully decentralized learning offers a promising solution that avoids any single point of vulnerability (as with federated learning).
\end{example}

\begin{example}\label{ex:iot}
As the prevalence of smart mobile devices generates vast amounts of data, central servers may face challenges in managing high latency, low bandwidth, and communication overhead when employing federated learning to learn general purpose models for object detection, natural language processing, recommendation systems, etc. Consequently, we envision a rising need for data private decentralized learning where training can be distributed across multiple servers to address these issues gracefully.
\end{example}

\noindent We motivate our framework's design to protect data privacy for applications like these by combining elements of decentralized optimization, differential privacy, and deep learning as described in the following sections.

\input{figures/hero}
\begin{figure}
  \centering
  \hero
  \caption{Illustration of a sample collaborative classification task. Drones with limited individual fields of view gather local observations of a global industrial landscape~\cite{DBLP:journals/corr/abs-1709-00029} below, denoted $\mathcal{D}_i$ for agent $i$. Neighboring drones communicate, as indicated by the waved lines, to learn joint classification from collective experiences. Image inspired by~\cite{mitchell2020gaussian}.}
  \label{fig:hero}
\end{figure}

\section{Detailed Convergence Analysis}

We provide more detailed proofs of the convergence analysis, written in abbreviated form in Section \ref{sec:conv_anal}.

\begin{theorem}[Linear Convergence Theorem]
Assume that $M(\theta)$ of the base decentralized algorithm is Lipschitz w.r.t. a constant $C_L$, i.e., $\|M(\theta) - M(\hat{\theta})\|_2 \le C_L \|\theta - \hat{\theta}\|_2$ for any $\theta$ and $\hat{\theta}$. If $M(\theta)$ of the base decentralized algorithm converges to the optimal solution $M(\theta^*)$ at a linear rate, then $M(\theta)$ of the differentially private decentralized algorithm also converges to the optimal solution $M(\theta^*)$ at a linear rate within an error neighborhood on the order of $\mathcal{O}(\sigma)$.
\label{thm:lin_convergence2}
\end{theorem}

\noindent Before proving Theorem \ref{thm:lin_convergence2}, note that since $M(\theta)$ of the base decentralized algorithm converges linearly, there exists $\rho \in (0,1)$ (the contraction factor) such that
\begin{align}\label{eq:1002}
    \| M\big(\theta^{(k+1)}\big) - M(\theta^*) \| \le \rho \| M(\theta^{(k)}) - M(\theta^*)\|
\end{align}
for all $k \ge 0$, where $\theta^*$ is the 
optimal solution and $\theta^{(k)}$ denotes the parameters at iteration $k$. The following lemma will be useful in proving Theorem \ref{thm:lin_convergence2}.

\begin{lemma}\label{lemma:12}
Let $\tilde{\theta}^{(k)}$ be the noisy parameters of the differentially private decentralized algorithm at iteration $k$, $\theta^{(0)}$ be the initial parameters, $\xi^{(k)}$ be the sampled noise at iteration $k$, $C_L$ be the Lipschitz constant of $M(\cdot)$ with respect to $\theta$, and $C_\sigma$ be a scaling constant on the noise that depends on the specific update step $\Psi(\cdot)$ of the base algorithm. Then, it holds that
\begin{align}\label{eq42}
    \| M\big(\tilde{\theta}^{(k)}) - M(\theta^*) \| &\le \rho^{k} \| M\big(\theta^{(0)}\big) - M(\theta^*) \| \\
    &+ C_L C_\sigma \sum_{i=0}^{k} \rho^i \eta^{(k-i)} \|\xi^{(i)}\|. \nonumber
\end{align}
\end{lemma}

\proof{
We use induction to prove the lemma. For the initial iteration $k=0$, we use the triangle inequality to observe
\begin{align}\label{eq52}
    \| M\big(\tilde{\theta}^{(0)}) - M(\theta^*) \|
    &\le \| M\big(\tilde{\theta}^{(0)}) - M(\theta^{(0)}) \| \\
    &+ \| M\big(\theta^{(0)}\big) - M(\theta^*) \| \nonumber
\end{align}
Considering the $\|M(\tilde{\theta}^{(0)}) - M(\theta^{(0)}) \|$ term first, we can apply the Lipschitz property of $M(\cdot)$ and simplify using the parameter update rule $\tilde{\theta}^{(k)} = \theta^{(k)} + \eta^{(k)} C_\sigma \xi^{(k)}$ to find
\begin{align}\label{eq62}
    \|M(\tilde{\theta}^{(0)}) - M(\theta^{(0)}) \| &\le C_L \| \tilde{\theta}^{(0)} - \theta^{(0)} \| \\
    &= C_L \eta^{(0)} C_\sigma \| \xi^{(0)} \| \nonumber.
\end{align}
Substituting \eqref{eq62} into \eqref{eq52} shows that \eqref{eq42} holds for $k=0$. For the inductive case, we assume \eqref{eq42} holds for iteration $k \ge 0$ and consider iteration $k+1$. From \eqref{eq:1002} and updating $\tilde{\theta}^{(k)}$ with the base decentralized algorithm, we have
\begin{align}\label{eq102}
    \| M\big(\theta^{(k+1)}\big) - M(\theta^*) \|
    &\le \rho \| M(\tilde{\theta}^{(k)}) - M(\theta^*)\|.
\end{align}
By substituting \eqref{eq42} into \eqref{eq102}, we obtain
\begin{align}\label{eq132}
    &\| M\big(\theta^{(k+1)}\big) - M(\theta^*) \| \\
    &\le \rho^{k+1} \| M\big(\theta^{(0)}\big) - M(\theta^*) \| + C_L C_\sigma \sum_{i=1}^{k+1} \rho^i \eta^{(k+1-i)} \| \xi^{(i)} \|.\nonumber
\end{align}
Furthermore, by the triangle inequality and Lipschitz condition of $M(\cdot)$, we have
\begin{align}\label{eq142}
    &\| M\big(\tilde{\theta}^{(k+1)}) - M(\theta^*) \| \\
    &\le \| M\big(\tilde{\theta}^{(k+1)}) - M(\theta^{(k+1)}) \| + \| M\big(\theta^{(k+1)}\big) - M(\theta^*) \| \nonumber \\
    &= \| M\big(\theta^{(k+1)}\big) - M(\theta^*) \| + C_L \eta^{(k+1)} C_\sigma \| \xi^{(k+1)} \|.\nonumber
\end{align}
By substituting \eqref{eq132} into \eqref{eq142}, we find the desired result
\begin{align}\label{eq152}
    \| M\big(\tilde{\theta}^{(k+1)}\big) - M(\theta^*) \| &\le \rho^{k+1} \| M\big(\theta^{(0)}\big) - M(\theta^*) \| \\
    &+ C_L C_\sigma \sum_{i=0}^{k+1} \rho^i \eta^{(k+1-i)} \xi^{(i)}. \nonumber
\end{align}
Thus, we have shown that \eqref{eq42} holds for $k+1$.
}

\noindent Now we are equipped to prove Theorem \ref{thm:lin_convergence2}.

\proof{
Since each noise term $\xi^{(k)}$ is independently drawn from a normal Gaussian distribution, we can take the expectation over the result from Lemma \ref{lemma:12} to get
\begin{align}\label{eq:172}
    \mathbb{E} \big[ \| M\big(\tilde{\theta}^{(k)}) - M(\theta^*) \| \big] &\le \rho^{k} \mathbb{E} \big[ \| M\big(\theta^{(0)}\big) - M(\theta^*) \| \big] \\
    &+ C_L C_\sigma \sum_{i=0}^{k} \rho^i \eta^{(k-i)} D \nonumber
\end{align}
where $D = \mathbb{E} \big[ \|\xi\| \big] = \mathcal{O}(\sigma)$ since $\xi$ 
has a standard deviation related to $\sigma$. Since $\sum_{i=0}^k \rho^i$ is the sum of the geometric series with $\rho \in (0,1)$ and the learning rate is bounded for all iterations, we have
\begin{align}\label{eq182}
    \sum_{i=0}^k \rho^i \le \frac{1}{1-\rho}~\text{and}~\eta^{(k)} \le C_\eta, ~\forall~k \ge 0.
\end{align}
By substituting \eqref{eq182} into \eqref{eq:172}, we find
\begin{align}\label{eq192}
     &\mathbb{E} \big[ \| M\big(\tilde{\theta}^{(k)}) - M(\theta^*) \| \big] \\
     & \le \rho^k \mathbb{E} \big[ \| M\big(\theta^{(0)}\big) - M(\theta^*) \| \big] + \frac{C_L C_\sigma C_\eta }{1-\rho} D. \nonumber
\end{align}
Therefore, we complete the proof with
\begin{align}\label{eq202}
     \lim_{k \to \infty}\mathbb{E} \big[ \| M\big(\tilde{\theta}^k) - M(\theta^*) \| \big] \le \mathcal{O}(\sigma).
\end{align}
Note that we do not consider gradient clipping in the above analysis because gradient clipping is equivalent to reducing the step size $\eta^k$, which may rescale the contraction factor $\rho$ of the linear rate but does not affect the convergence behavior.
}

\noindent The base algorithms DSGD, DSGT, and DiNNO converge linearly when the objective function is strongly convex~\cite{NIPS2017_a74c3bae,pu2020distributed,DBLP:journals/corr/abs-2109-08665}. For other possible cases, the following theorem generalizes our convergence analysis with additional assumptions.

\begin{theorem}[General Convergence Theorem]
Assume the gradient function $\nabla l(\cdot)$ is Lipschitz w.r.t the variable $\theta$, the update step of the base algorithm $\Psi(\cdot)$ is Lipschitz w.r.t. the gradient $G$ and normalized Lipschitz w.r.t. the variable $\theta$, the step size satisfies $\eta^k \le \frac{\hat{\eta}^k}{C k}$ at each iteration $k$ (where $\hat{\eta}^k \le 1$ is any sequence of values with a finite sum $\sum_{k=1}^\infty \hat{\eta}^k \le \infty$), and $C$ is a constant depending on the Lipschitz properties of $\nabla l(\cdot)$ and $\Psi$. If $\theta$ of the base decentralized algorithm converges to the optimal solution $\theta^*$, then $M(\theta)$ of the differentially private decentralized algorithm also converges to the optimal solution $M(\theta^*)$ in an error neighborhood of $\mathcal{O}(\sigma)$.
\label{thm:general_convergence2}
\end{theorem}

\noindent Let $\tilde{\theta}^{(k)}$ be the differentially private algorithm's variable vector and $\theta^{(k)}$ be the base algorithm's variable vector on the kth iteration. Note that from Algorithm \ref{alg:framework} that $\tilde{\theta}^{(k)}$ deviates from 
$\theta^{(k)}$ 
with the accumulated noise from previous iterations. We can upper bound the difference between these decision variables as 
\begin{align}\label{eq:02}
\mathbb{E}[\|\tilde{\theta}^{(k+1)} - \theta^{(k+1)}\|] &\le C_{\sigma} \eta^{k+1} \big(\mathbb{E}[\|\tilde{\theta}^{(k)} - \theta^{(k)}\|]\\
&+ \mathbb{E}[\|\xi^{(k+1)}\|] \big) + \mathbb{E}[\|\tilde{\theta}^{(k)} - \theta^{(k)}\|] \nonumber
\end{align}
where $C_{\sigma}$ is a scaling constant for the noise. The first term in the bound results from the gradient deviation and $C_{\sigma}$ depends on the Lipschitz constants of the gradient function and the update step of the base algorithm, while the second term results from the variable deviation. With this observation, we can use induction to prove the following useful lemma

\begin{lemma} \label{lem:general2}
On each iteration $k$, we can bound the expected difference between $\tilde{\theta}^{(k)}$ and $\theta^{(k)}$ with
\begin{equation}\label{eq:12}
\mathbb{E}[\|\tilde{\theta}^{(k)} - \theta^{(k)}\|] \le \sum_{\kappa=1}^{k} \hat{\eta}^k \sigma.
\end{equation}
\end{lemma}

\proof{
By definition, we initially have $\tilde{\theta}^{(0)} = \theta^{(0)}$. For iteration $k=1$, substituting this into \eqref{eq:02} gives
\begin{equation}
\mathbb{E}[\|\tilde{\theta}^{(1)} - \theta^{(1)}\|] \le C_{\sigma} \eta^{1} \mathbb{E}[\|\xi^{(1)}\|]. 
\end{equation} 
By letting $C \ge C_{\sigma}$ and using the fact $\eta^1 \le \frac{\hat{\eta}^1}{C}$, we get
\begin{equation}
\mathbb{E}[\|\tilde{\theta}^{(1)} - \theta^{(1)}\|] \le \hat{\eta}^1 \mathbb{E}[\|\xi^{(1)}\|] \le \hat{\eta}^1 \sigma, 
\end{equation}
and thus \eqref{eq:12} holds for iteration $k=1$. For the inductive case, we assume \eqref{eq:12} holds for iteration $k \geq 1$ and consider iteration $k+1$. By substituting \eqref{eq:12} into \eqref{eq:02}, we have
\begin{equation}\label{eq:22}
\mathbb{E}[\|\tilde{\theta}^{(k+1)} - \theta^{(k+1)}\|] \le C_{\sigma} \eta^{k+1} \Big(\sum_{\kappa=1}^{k} \hat{\eta}^k \sigma + \sigma\Big) + \sum_{\kappa=1}^{k} \hat{\eta}^k \sigma, \nonumber
\end{equation}
and using $\eta^k \le \frac{\hat{\eta}^k}{C k} \le \frac{\hat{\eta}^k}{C_\sigma k}$ and $\hat{\eta}^k \le 1$,
we find
\begin{align*}
\mathbb{E}[\|\tilde{\theta}^{(k+1)} - \theta^{(k+1)}\|] &\le \frac{\hat{\eta}^{k+1}}{k+1} \sum_{\kappa=1}^{k+1} \sigma + \sum_{\kappa=1}^{k} \hat{\eta}^{\kappa} \sigma \le \sum_{\kappa=1}^{k+1} \hat{\eta}^{\kappa} \sigma.
\end{align*}
Thus, \eqref{eq:12} holds for iteration $k+1$, completing the proof.
}

\noindent Next we use Lemma \ref{lem:general2} to finish proving Theorem \ref{thm:general_convergence2}.

\proof{
Using \eqref{eq:12} from Lemma \ref{lem:general2} and the condition that there exists $C_\eta$ such that $\sum_{k=1}^\infty \hat{\eta}^k \le C_\eta > 0$, we get
\begin{equation}
\mathbb{E}[\|\tilde{\theta}^{(k)} - \theta^{(k)}\|] \le C_\eta \sigma, \nonumber
\end{equation}
where $\sigma$ is the standard deviation of the Gaussian sample. Therefore, we conclude the proof that 
\begin{equation}
\lim_{k \to \infty} \mathbb{E}[\|M(\tilde{\theta}^{(k+1)} - M(\theta^{(k+1)}\|] \le C_L C_\eta \sigma. \nonumber
\end{equation} 
Note that the exact expression of the scaling constant $C_\sigma$ in our analysis depends on the base algorithm's update step.
}

\section{Additional Evaluations}

\subsection{Graph Density and Sparsity}

In this section, we investigate the impact of the communication graph topology on consensus rates and performance for decentralized algorithms under differential privacy by varying the graph density/sparsity. This is in a similar vein to Section \ref{sec:alg_con} where we study the same impact by varying the graph algebraic connectivity. We use the common understanding of density in undirected graphs in Definition 7.2.
\smallbreak
\noindent \textbf{Definition 7.2} (Density). 
Let $N$ be the number of agents in the graph and $M$ be the number of communication links or edges between them. The density $d$ of the graph is defined as:
\begin{equation}
d = \frac{2M}{N(N-1)}
\end{equation}

\begin{figure}[h]
  \centering
  \includegraphics[width=\linewidth]{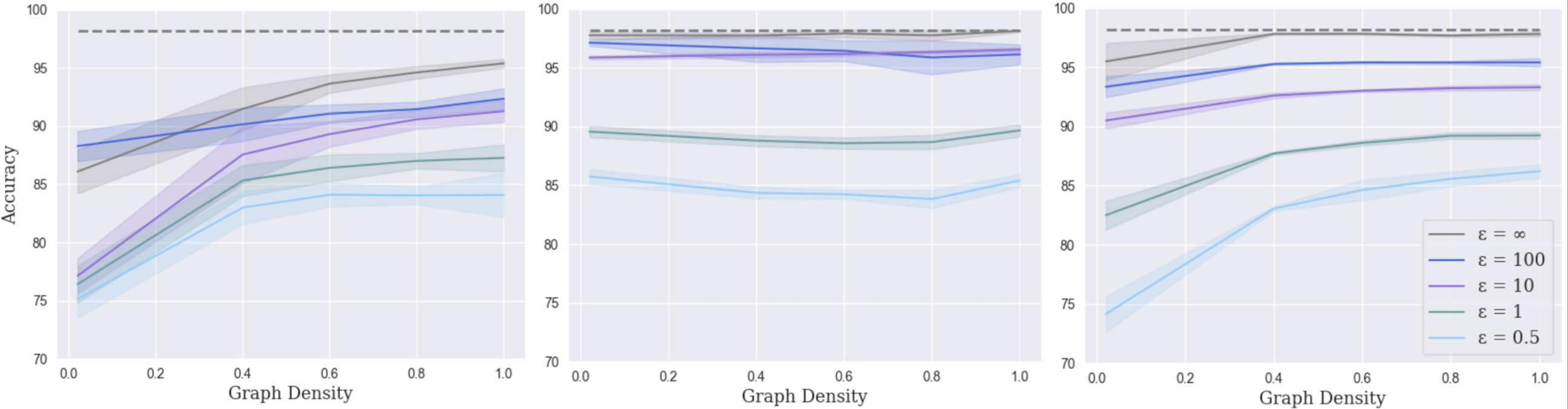}
  \caption{All DP decentralized algorithms are relatively invariant to connectivity measured by density up to a threshold value near 0.4. The non-private accuracy from single-agent SGD is denoted with a dashed line.}
 \label{fig:density}
\end{figure}

We choose to vary density since this interpretable metric characterizes how close the number of edges in the graph is to the maximal number of edges and directly correlates with dissemination of inter-agent information. We train DP-DSGD, DP-DSGT, and DP-DiNNO on distributed MNIST with $t=1$ using randomly generated communication graphs of density values close to one of five selected values (0.2, 0.4, 0.6, 0.8, and 1). We run five trials for each datapoint and show how accuracy changes with graph density and privacy budgets in Figure \ref{fig:connectivities} (bottom row). These trends are largely consistent with our findings when varying the algebraic connectivity. Specifically, we see that DP-DSGT is invariant to extremely sparse graphs and DP-DSGD and DP-DiNNO are invariant to sparsity up to a threshold.

\subsection{Node Centrality}

\begin{figure}[h]
  \centering
  \includegraphics[width=\linewidth]{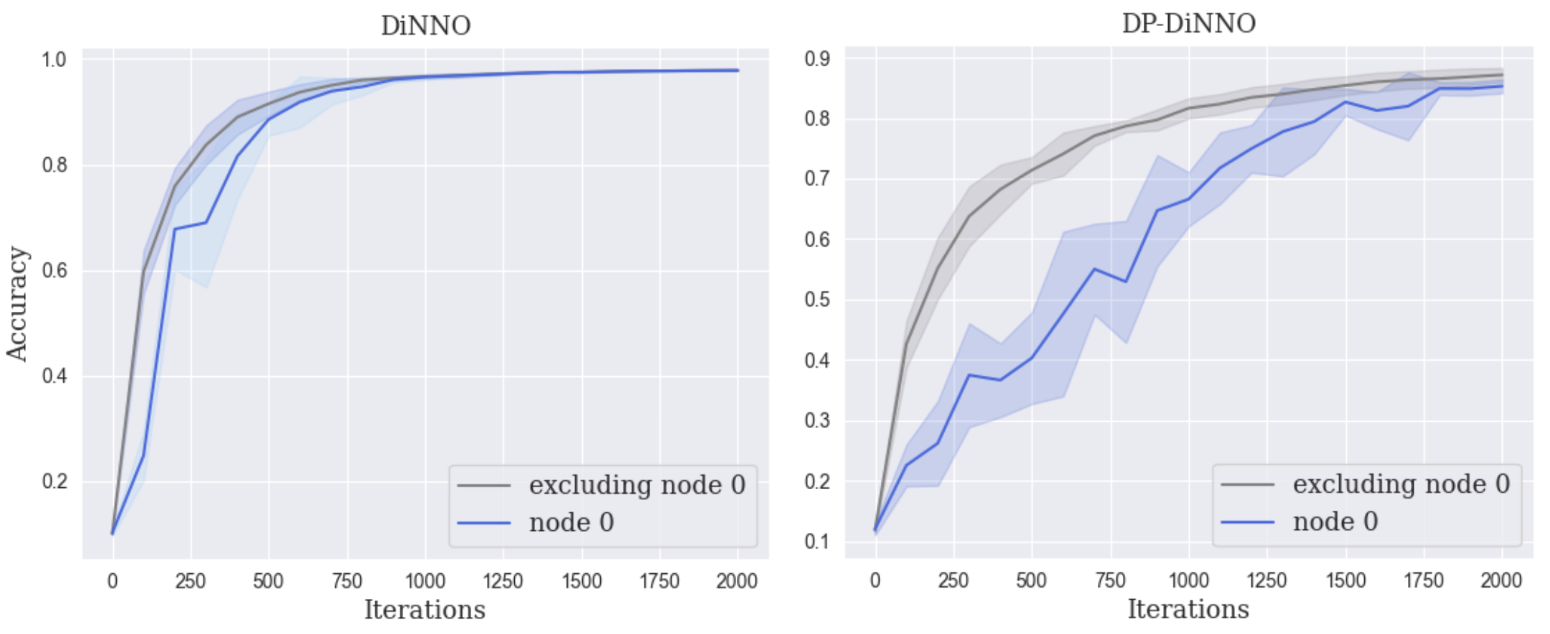}
  \caption{Node 0's validation accuracy converges to the system average faster when training with DiNNO (top) than with $(1,10^{-5})$-DP-DiNNO (bottom). For both evaluations, node 0 has 0.05 eigenvector centrality.}
 \label{fig:dinno_node_centrality}
\end{figure}

\begin{figure}[h]
  \centering
  \includegraphics[width=\linewidth]{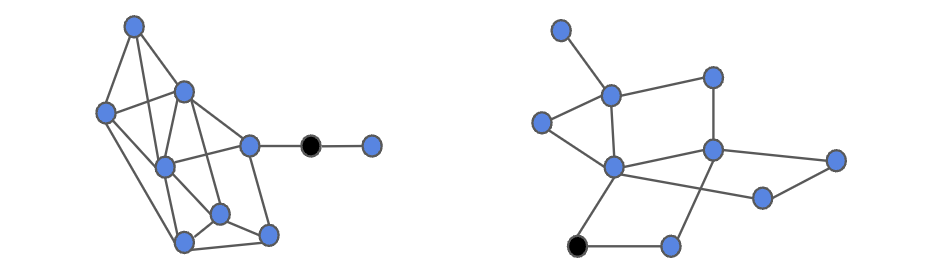}
  \caption{Example communication graphs used for decentralized training. The left graph has algebraic connectivity 0.38, density 0.4, and the black node has eigenvector centrality 0.05. The right graph has algebraic connectivity 0.62, density 0.29, and the black node has eigenvector centrality of 0.5.}
 \label{fig:graphs}
\end{figure}

To demonstrate that non-private DiNNO is minimally affected by node centrality, as discussed in Section \ref{sec:node_centrality}, we visualize the training performance of node 0 vs other nodes in Figure \ref{fig:dinno_node_centrality}. This finding extends to DP-DSGD and DP-DSGT as well. Visualizations of example communication graphs are given in Figure \ref{fig:graphs}.

\end{document}

%% file: figures/hero.tex
\tdplotsetmaincoords{70}{20}
\newcommand{\cs}{*0.2}
\def\centerarc[#1](#2)(#3:#4:#5)
    { \draw[#1] ($(#2)+({#5*cos(#3)},{#5*sin(#3)})$) arc (#3:#4:#5); }

\tikzset{
  pics/drone/.style args={#1}{code={
    \draw (0, 0, 0) coordinate (o)
        -- ++(-#1,-#1,0) coordinate (a) -- cycle (o)
        -- ++(#1,-#1,0) coordinate (b) -- cycle (o)
        -- ++(-#1,#1,0) coordinate (c) -- cycle (o)
        -- ++(#1,#1,0) coordinate (d);
    \foreach \p in {(a), (b), (c), (d)}:
        \draw \p circle[radius=0.9*#1];
  }},
  pics/cross/.style args={#1}{code={
    \draw (0, 0, 0) coordinate (o)
        -- ++(-#1,0,0) -- cycle (o)
        -- ++(#1,0,0) -- cycle (o)
        -- ++(0,-#1,0) -- cycle (o)
        -- ++(0,#1,0);
  }},
  pics/agent/.style args={#1}{code={
      \draw (-0.3,0,0) coordinate (a) -- ++(0,\size,0) coordinate (b) -- ++(\size,0,0) coordinate (c) -- ++(0,-\size, 0) coordinate (d) -- (a);
      \draw (-0.3,0.3,0) coordinate (a) -- ++(0,\size,0) coordinate (b) -- ++(\size,0,0) coordinate (c) -- ++(0,-\size, 0) coordinate (d) -- (a);
      \draw (-0.3,0.6,0) coordinate (a) -- ++(0,\size,0) coordinate (b) -- ++(\size,0,0) coordinate (c) -- ++(0,-\size, 0) coordinate (d) -- (a);
      \draw (-0.6,0.6,0) coordinate (a) -- ++(0,\size,0) coordinate (b) -- ++(\size,0,0) coordinate (c) -- ++(0,-\size, 0) coordinate (d) -- (a);
      \draw (0,0,0) coordinate (a) -- ++(0,\size,0) coordinate (b) -- ++(\size,0,0) coordinate (c) -- ++(0,-\size, 0) coordinate (d) -- (a);
      \coordinate (o) at (0.5*\size, 0.5*\size, 2); 
      \coordinate (g) at (0.5*\size, 0.5*\size, 0); 
      \foreach \p in {(a), (b), (c), (d)}:
          \draw[dashed] \p -- (o);
      \node[above=0.1cm of o] (agent_i) {$#1$};
      \pic[gray] (drone) at (0.5*\size, 0.5*\size, 2) {drone={0.2}};
  }},
  size/.store in=\size,
  size=9\cs,
  comm/.style = {decorate,decoration={snake, amplitude=0.4mm, segment length=3mm}},
}

\newcommand{\hero}{
    \begin{tikzpicture}[tdplot_main_coords]
        \node[canvas is xy plane at z=0] (temp) at (0,0,0) {\includegraphics[width=6.4cm]{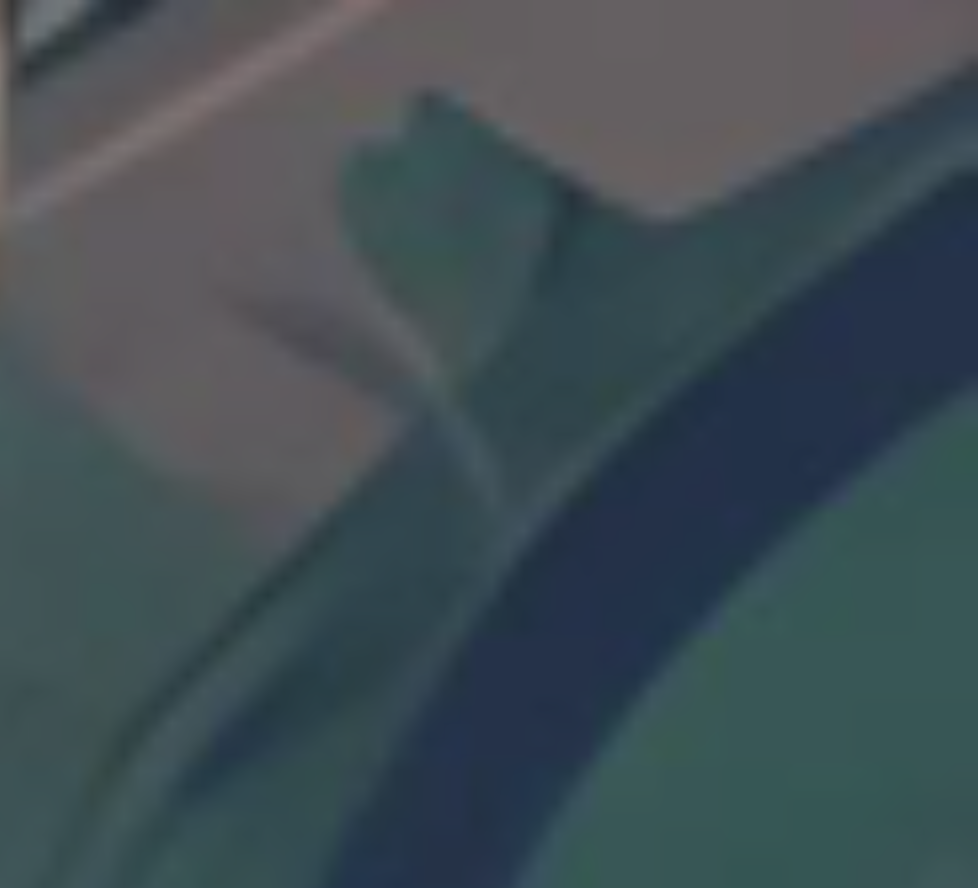}};
        \pic (aa) at (-13\cs, -13\cs, 0) {agent={a}};
        \pic (ai) at (-8\cs, 2\cs, 0) {agent={b}};
        \pic (ab) at (6\cs, -13\cs, 0) {agent={c}};
        \draw[comm] (aao) -- node [pos=0.5, above] {} (aio);
        \draw[comm] (abo) -- node [pos=0.45, above] {} (aio);

        
        
        \node[above right=0.2cm and 0.3cm of aio] {$\mathcal{N}_b = \{a, b, c\}$};
        \node[above left=-0.1cm of abg] {$\mathcal{D}_c$};
        \node[above left=-0.1cm of aig] {$\mathcal{D}_b$};
        \node[above left=-0.1cm of aag] {$\mathcal{D}_a$};
        
    \end{tikzpicture}
}